\documentclass[10pt, a4paper]{article}

\usepackage[final]{lrec-coling2024} %

\usepackage{hyperref}
\usepackage{xcolor}
\definecolor{darkblue}{rgb}{0, 0, 0.5}
\hypersetup{colorlinks=true, citecolor=darkblue, linkcolor=darkblue, urlcolor=darkblue}

\usepackage{makecell}
\usepackage{times}
\usepackage{epsfig}
\usepackage{graphicx}
\usepackage{amsmath}
\usepackage{amssymb}
\usepackage{amsfonts}

\usepackage{url}            %
\usepackage{booktabs}       %
\usepackage{nicefrac}       %
\usepackage{microtype}      %
\usepackage{tabularx}
\usepackage{float}
\usepackage{outlines}
\usepackage{algorithm}
\usepackage{algpseudocode}
\usepackage{chngcntr}
\usepackage{float}
\usepackage{dsfont}
\usepackage{subfigure}
\usepackage{etoolbox}
\usepackage[subtle,paragraphs=normal]{savetrees}
\usepackage{comment}
\usepackage{xcolor}

\newcommand{\eg}{\textit{eg.}}

\title{Multi-Stage Multi-Modal Pre-Training For Automatic Speech Recognition}

\name{Yash Jain$^{\dagger \mathsection 1}$, \quad David Chan$^{\ast \mathsection 1}$, \quad Pranav Dheram$^{\mathsection}$, \quad Aparna Khare$^{\mathsection}$, \\  {\bf \large Olabanji Shonibare$^{\mathsection}$, \quad Venkatesh Ravichandran$^{\mathsection}$, \quad Shalini Ghosh$^{\mathsection}$}} 

\address{$^{\dagger}$Georgia Institute of Technology\\
         $^{\ast}$University of California, Berkeley (EECS) \\
         $^{\mathsection}$Amazon Alexa AI \\
        yashjain@gatech.edu}
\abstract{
Recent advances in machine learning have demonstrated that multi-modal pre-training can improve automatic speech recognition (ASR) performance compared to randomly initialized models, even when models are fine-tuned on uni-modal tasks. Existing multi-modal pre-training methods for the ASR task have primarily focused on single-stage pre-training where a single unsupervised task is used for pre-training followed by fine-tuning on the downstream task. In this work, we introduce a novel method combining multi-modal and multi-task unsupervised pre-training with a translation-based supervised mid-training approach. We empirically demonstrate that such a multi-stage approach leads to relative word error rate (WER) improvements of up to 38.45\% over baselines on both Librispeech and SUPERB. Additionally, we share several important findings for choosing pre-training methods and datasets.
 \\ \newline  }

\begin{document}

\maketitleabstract
\footnotetext[1]{Work done during an internship at Amazon Alexa AI.}

\section{Introduction}

Despite progress in large-scale pre-training for automatic speech recognition (ASR) \cite{chen2022wavlm, hsu2022u, chan2022avbert}, uni-modal (speech-only) ASR remains a challenging task, particularly when faced with rare words and noisy acoustic conditions. When understanding spoken phonemes, the model must correctly discern both speaker-specific patterns (e.g., accent,  prosody) and global noise patterns (e.g., background noise, intermittent interruptions, confounding speakers). Recent work in natural language processing (NLP) \cite{tu2020empirical,hoffmann2022training}, robotics \cite{mandlekar2022matters, kuhar2023learning, khazatsky2024droid} and computer vision  \cite{goyal2022vision,NEURIPS2023_d1786f52, jain2024damex} has demonstrated that exposing models to a \textit{high diversity} of data during pre-training is essential in building robust representations.

Similarly, recent works in the ASR community have corroborated these results. \citet{shi2022learning} and \citet{hsu2022u} demonstrated that pre-training on large-scale audio-visual data (or audio-only data), in the form of lip-reading videos, leads to better performance on the lip-reading task. \citet{chan2022avbert} showed that exposing models to video data during pre-training led to performance improvements not only when visual input is available at training time, but also when \textit{only audio is available at test time}.

\citet{chan2022avbert} also demonstrated that adding visual information from non-speech specific videos (leveraging the Kinetics dataset \cite{carreira2017quo}) is only a small portion of the possible augmentations that can be made during pre-training. In this work, we not only explore two new audio-visual pre-training sources, but also leverage a translation task with English speech input as a new mid-training task to consolidate information learned during the pre-training phase. Further, while  \citet{chan2022avbert} explore an attention-based transfer-learning framework based on k-means clustering for pre-training, we simplify the pre-training architecture significantly, and explore several pre-training objectives beyond masked cluster prediction. Our primary contributions are as follows:

\begin{enumerate}
\item We perform large-scale evaluation of multiple audio-visual pre-training methods (MAE, CLR) using several pre-training datasets (Kinetics, VoxCeleb2, LRS3) with varying characteristics. We evaluate them on the ASR task and the SUPERB benchmark, showing how  multi-modal pre-training is affected by key dataset characteristics.
\item We show that pre-training with audio-visual data, particularly data from speech-specific audio-visual datasets can improve word error rate (WER) up to 30.8\% relative compared to randomly initialized baseline models on speech-only test data.
\item We introduce a novel mid-training stage between the pre-training and fine-tuning steps, using speech translation as the mid-training task. The mid-training stage improves WER by 38.45\% relative on the Librispeech test-clean dataset, and by 26.18\% relative on the test-other dataset compared to audio-visual pre-training only baseline. The technique also shows improvements on several tasks (Keyword Spotting, Intent Classification, Phoneme Recognition, and Speaker Diarization) in the SUPERB \cite{yang2021superb} benchmark.
\end{enumerate}

\section{Background}

Representation learning methods like Contrastive Predictive Coding \cite{oord2018representation} and Wav2Vec \cite{schneider2019wav2vec} have shown significant promise when applied to ASR. Methods for large-scale pre-training for ASR can be categorized into two methods: masked autoencoding methods \cite{hsu2021hubert, chen2022wavlm}, and contrastive learning \cite{baevski2020wav2vec}. While traditionally self-supervised methods are trained on a single target loss, other methods have been proposed which leverage multiple pre-training targets. \citet{pascual2019learning,talnikar2021joint,wang2021unispeech} all optimize a combination of uni-modal supervised losses and recently, approaches such as  W2v-BERT \cite{chung2021w2v} and JUST \cite{bai2022joint} have combined contrastive approaches with masked auto-encoding to build robust self-supervised speech representations. Similarly, while most self-supervised methods are pre-trained on a single dataset, \citet{radford2022robust,narayanan2018toward,likhomanenko2020rethinking,chan2021speechstew} have all demonstrated that a wide mix of data is essential for pre-training. In this work, we target both of these problems: use a combination of losses, and pre-training stages under different datasets to improve the learned multi-modal representations.

Audio-visual data provides diverse information for representation learning. \citet{shi2022learning} demonstrate improvements on ASR when visual input is available (at both training and test time), and methods such as u-HuBERT \cite{hsu2022u} extend such pre-training approaches to applications where both uni-modal and multi-modal data are available at training-time (but still require multi-modal data for inference). Later work by \citet{chan2022avbert} demonstrated that pre-training with paired audio-visual data, can even improve performance on \textit{uni-modal} datasets.

In addition to multiple modalities, pre-training with multiple languages has also been explored in the literature.  \citet{radford2022robust} demonstrate that pre-training with a wide range of inputs from several languages improves ASR performance across all of the studied languages. \cite{lahiri2021multilingual} show that leveraging self-supervised learning (SSL) for knowledge transfer across languages can yield WER improvements of up to 3.55\% relative WER on target languages, and \citet{karimi2022deploying} demonstrate that in almost all cases, even out-of-domain multi-lingual data can improve WER in single and multi-speaker conversations and dictation tasks. 

\begin{figure*}[!ht]
    \centering
    \includegraphics[width=1\linewidth,trim=100 120 100 120, clip]{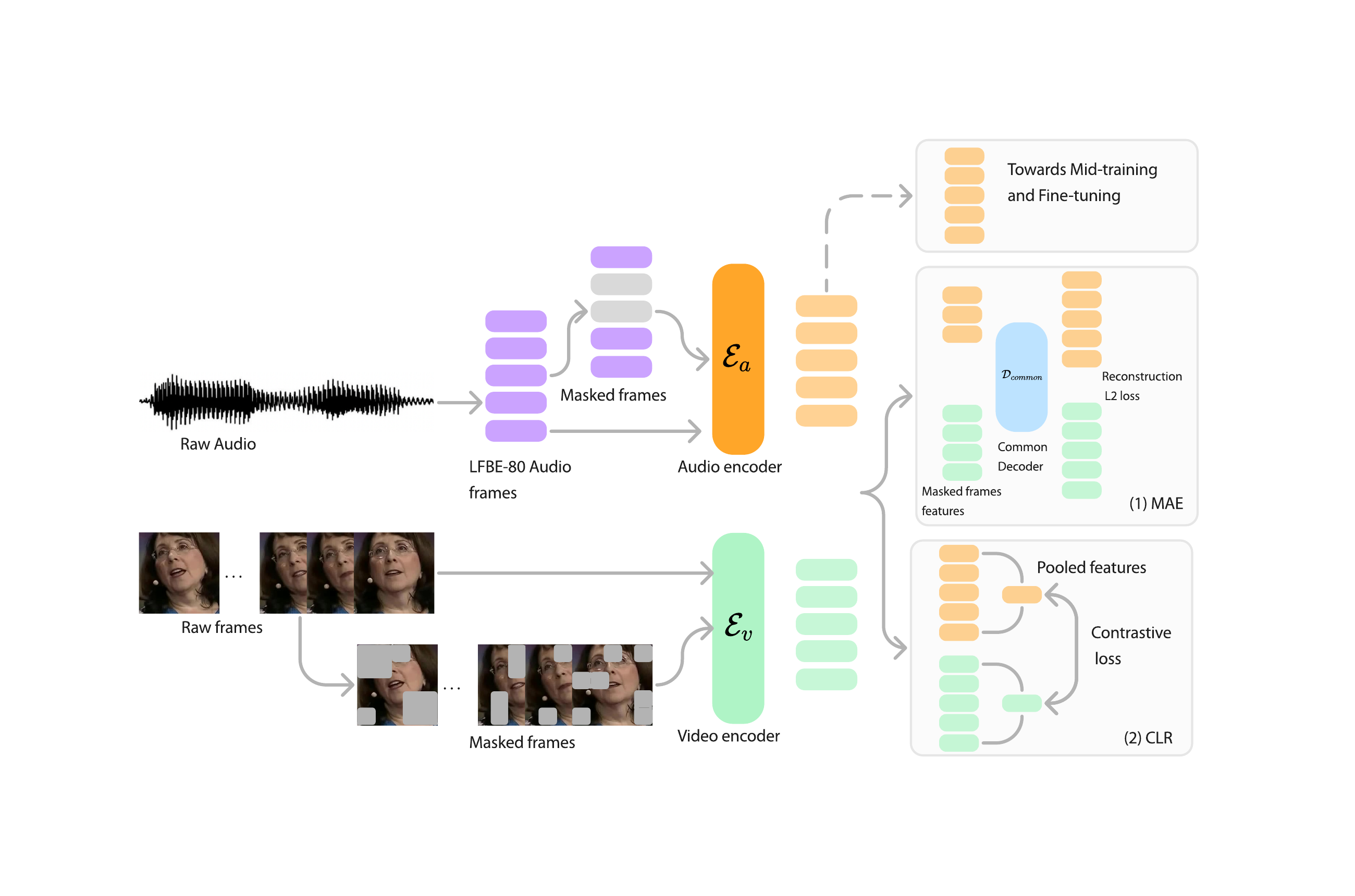}
    \caption{\small \textbf{Overview of multi-modal training strategy}. Raw audio and video features are extracted from source data. These features are then passed through the audio and video encoders to get features which are further processed as (1) MAE: the masked encoded features are reconstructed through a common decoder successively and are compared against original input using L2 loss, (2) CLR: contrastive learning applied to spatio-temporally pooled audio and video encoded features, and (3) the trained audio encoder is further used for mid-training (translation task) and then for downstream tasks.}
    \label{fig:method}
\end{figure*}

\section{Methods}

Our method (\autoref{fig:method}), consists of a multi-stage multi-modal pre-training approach, followed by a fine-tuning stage on downstream tasks. We describe our method in this section.

\subsection{Pre-Training Tasks}

We experiment with two pre-training strategies that differ in the granularity of information they extract. The first method, Masked Autoencoder (MAE), learns local features by reconstructing masked parts of speech and video. The second method, Contrastive Learning (CLR),  focuses on global features by using pooled audio-visual features from the same video as positive pairs while other combinations of audio-visual pairs as negatives. The two pre-training strategies help us compare the effects of local and global feature learning against the visual-audio dataset characteristics, for \eg, Kinetics dataset \cite{k600} has non-speech audio streams, while LRS-3 \cite{afouras2018lrs3} and Voxceleb2 \cite{chung2018voxceleb2} datasets have videos with speech. \\

\noindent \textbf{Masked Autoencoding (MAE):} Traditional MAE approaches for ASR pre-training have focused on token-based reconstruction \cite{hsu2021hubert, shi2022learning, chan2022avbert}. However these methods have the drawback of requiring a separate quantization method, which can add significant training complexity. We simplify the encoder to directly reconstruct features from the original masked signal. 

Our MAE approach consists of three encoders: $\mathcal{E}_a$, a masked audio-specific encoder based on the encoder in \citet{chen2022wavlm}, $\mathcal{E}_v$, a masked video-specific encoder based on \citet{tong2022videomae}, and $\mathcal{D}_{a+v}$, a joint transformer decoder with the same structure as in \citet{devlin2018bert}.

Let $a \in \mathbb{R}^{T_a \times F}$ be the set of audio input frames (we use f-dimensional log-filterbank energies (LFBE)), and $v \in \mathbb{R}^{H // P_H \times W // P_W \times T_v // P_T \times (P_HP_WP_T)}$ be a set of video frames, which have been subdivided into $(P_H, P_W, P_T)$ voxels. $T_a$ refers to number of audio frames and $T_v$ are number of video frames of height $H$ and width $W$. To generate the input sequence to $\mathcal{E}_a$, we randomly mask a fraction $\phi$ of the audio frames with 0s (masking), and generate the embedded audio $e_a = \mathcal{E}_a(a)$. We use a similar process to mask voxels, to generate $e_v = \mathcal{E}_v(v)$.

The encoded representations $e_a$ and  $e_v$ are passed through the common decoder $\mathcal{D}_{a+v}$ to produce $o_a = \mathcal{D}_{a+v}(e_a)$ and $o_v = \mathcal{D}_{a+v}(e_v)$ respectively. The common decoder $\mathcal{D}_{a+v}$ ensures that the representations  $e_a$ and $e_v$ are projected to the same representation space.
The final MAE loss is computed as the squared L2 distance between $o_a$ and $a$, and $o_v$ and $v$:
\begin{equation}
\mathcal{L}_{\text{MAE}} = || \mathcal{D}_{a+v}(\mathcal{E}_a(a)) - a||_2^2 + || \mathcal{D}_{a+v}(\mathcal{E}_v(v)) - v||_2^2
\end{equation}
\\

\noindent \textbf{Contrastive Learning (CLR):} Contrastive Learning aims to learn representations using a contrastive loss that minimizes the distance between similar points and maximizes the distance between dissimilar points in a latent space. For contrastive learning, following \citet{radford2021clip} and \citet{xu2021videoclip}, we use the modality specific encodings $e_a$ and $e_v$ to generate $a^{enc} = \text{Pool}(e_a)$, where the pooling operation is a temporal average, and $v^{enc} = \text{Pool}(e_v)$, where the pooling operation is a spatio-temporal average. While other pooling operations like attention pooling are possible, we found that the spatio-temporal average captures consistent low-frequency global information, which correlates well with the information shared with the visual modality (unlike high-frequency information, which is often not evident from the visual modality).  The self-supervised contrastive loss for a batch of samples $a^{enc}_i, 1 \le i \le N$, and  $v^{enc}_i, 1 \le i \le N$ is computed as

\begin{equation}
\mathcal{L}_{\text{contrastive}}^{i} = -\log \left( \frac{\exp(a^{enc}_i \cdot v^{enc}_i)}{\sum_{k=1}^{N} \mathds{1}_{[k \neq i]} \exp(a^{enc}_i \cdot v^{enc}_k)} \right)
\end{equation}

\begin{equation}
\mathcal{L}_{\text{CLR}} = \frac{1}{N} \sum_{i=1}^N \mathcal{L}_{\text{contrastive}}^{i}
\end{equation}

\noindent \textbf{MAE + CLR:} In this setup, we combine the benefits of learning local features using MAE with learning global features using CLR as shown in \autoref{fig:method}. Both pre-training losses are added with equal weights, similar to \citet{chung2021w2v} to compute the final loss as

\begin{equation}
    \mathcal{L}_{\text{MAE+CLR}} = \frac{\mathcal{L}_{\text{MAE}} + \mathcal{L}_{\text{CLR}} }{2}
\end{equation}

\noindent\textbf{Pre-training Datasets:} We use three datasets for pre-training. The Kinetics-600 dataset \cite{k600} has 966 hours of audio-visual data for activity recognition, with a focus on the environment or instrument used. The videos contains non-speech audio data and have been used previously for audio-visual training \cite{chan2022avbert}. Voxceleb2 \cite{chung2018voxceleb2} provides 2380 hours of multi-lingual speaker recognition data with challenging acoustics and comprehensive lip and facial movements. LRS3 \cite{afouras2018lrs3} features 346 hours of clean, multi-modal spoken sentence data from TED and TEDx videos. The speech data in Voxceleb2 is has noisy acoustic conditions whereas LRS-3 has clean speech  with speakers talking to a close-talk microphone. These datasets allow for exploring the impact of clean-speech/noisy-speech/non-speech videos and pre-training techniques on the ASR task (\autoref{sec:ft}).

\subsection{Mid-Training: Speech Translation}

To improve performance of the pre-trained audio-visual models on the downstream tasks, we introduce a mid-training task that bridges the gap between pre-training and fine-tuning. Our approach transfers the learned distribution of the pre-trained model towards the distribution required for the downstream task, while discarding irrelevant information.

The mid-training task is designed to provide a low-cost warm-up for the pre-trained model, which can accurately represent various characteristics of the data. We chose to mid-train our audio encoder on the speech translation task using the MuST-C dataset \cite{di-gangi-etal-2019-must} in three languages, German, Italian and Dutch. This stage is useful for aligning the learned speech representations with the text modality which is beneficial for ASR, as shown in recent work in the speech representation learning space\cite{zhang2023google}. Our audio encoder was mid-trained until convergence on the speech translation task. This mid-training approach is the key to strong performance in downstream tasks, which we demonstrate in detail in \autoref{sec:results}.

Using translation as a mid-training task is only one possible instantiation of the mid-training approach. In addition to translation, future work can explore other speech-centric tasks like speaker identification, implied by \cite{chan2022contentcontext}), speaker/source separation, text to speech, and others. While we found that translation is effective in this work, we expect that each additional task will impact the downstream training process in unique ways.

\subsection{Fine-Tuning}
\label{sec:ft}
 
We evaluated our models by testing their performance on several downstream tasks. The fine-tuning task is distinct from the pre-training task of masked reconstruction (MAE) or contrastive learning (CLR), and the mid-training task designed to bridge the gap. Primarily, we evaluate the performance of the models on the test-clean and test-other Librispeech \cite{panayotov2015librispeech} datasets for ASR, as well as four tasks from the SUPERB \cite{yang2021superb} benchmark: Intent Classification (IC), Keyword Spotting (KS), Phoneme Recognition (PR) and Speaker Diarization (SD). Because our aim was to evaluate how both the pre-training and mid-training data distributions impact the final learned representations, we freeze the encoder weights during task specific fine-tuning, and fine-tune only the task specific decoder using the LS-960 dataset (for ASR) following \citet{baevski2020wav2vec} or the default datasets specified in the SUPERB benchmark \cite{yang2021superb}.

\subsection{Model Details}

In this section, we discuss the implementation details of the different training setups across the three datasets.
\\

\noindent \textbf{Video Data Pre-processing:} Videos are first resized to a resolution of $224 \times 224$ pixels, with a temporal stride of $4$ and $16$ frames sampled temporally. We apply random resized cropping with scale from $0.5$ to $1$, and random horizontal flipping following standard computer vision techniques for visual data augmentation. \\

\noindent \textbf{Video Encoder:} Our video encoding approach is similar to that of \cite{feichtenhofer2022masked}. Firstly, we divide the video into a regular grid of space-time patches of dimensions $16 \times 16 \times 2$ in the $(H, W, T)$ direction, respectively. These patches are then flattened and augmented with spatio-temporal positional embeddings \cite{vaswani2017attention}.

For the Masked Autoencoder, we randomly select 60\% of the patches for masking, and mask patches without replacement, while keeping the selection agnostic in the space-time domain. The remaining patches are then passed through 12 ViT encoder blocks \cite{dosovitskiy2020vit} with a hidden dimension of 768. We obtain the video encoded features of the remaining spatio-temporal patches, which are later reconstructed using a common decoder. 

For Contrastive Learning, we reduce the spatial patches to a single embedding for each frame \cite{xu2021videoclip, radford2021clip}. The reduced patches are  passed through a video encoder with 12 ViT encoder blocks \cite{dosovitskiy2020vit} with a hidden dimension of 768. The encoded embeddings are temporally pooled following \cite{xu2021videoclip}, resulting in single-vector video features which can be contrasted against corresponding audio embeddings. \\

\noindent \textbf{Audio Data Pre-processing:} The audio input is re-sampled to a frequency of 16kHz. Subsequently, 80-dimensional Log-Filterbank Energy (LFBE) features are computed from the resulting audio frames. To ensure consistency in feature size, we selected the first 1000 LFBE
frames for downstream processing. The frames are further sub-sampled using a 1D convolutional layer, reducing the number of audio frames to 250, following the approach of \citet{gulati2020conformer}. \\

\noindent \textbf{Audio Encoder:} We use positional embeddings in the sub-sampled audio frames similar to video encoding, as proposed by \citet{vaswani2017attention}. In the Masked Autoencoder, a random mask without replacement is applied to $60\%$ of the frames, with the visual and audio modalities sharing the same masking ratio to maintain balance in the amount of information across both modalities. The remaining frames are encoded by a Conformer \cite{gulati2020conformer} with 16 layers, 4 heads, and a depth-wise convolutional kernel of size 31. Audio features are then up-sampled by a linear layer 
and normalized for reconstruction.

In Contrastive Learning, the sub-sampled frames are directly featurized by the Conformer blocks without any masking involved. The audio features are then temporally pooled to obtain a single feature for the audio clip, which is up-sampled and normalized. For both the Mid-training and Fine-tuning tasks, the feature output from Conformer blocks is used as input to task-specific decoders. The weights of the convolutional sub-sampling layer and Conformer blocks are the only components re-used from the pre-training stage for further steps. \\

\noindent \textbf{Common Decoder:} The Masked Autoencoder pre-training step uses a relatively small vanilla ViT \cite{dosovitskiy2020vit} decoder of hidden dimension size of 512 and 4 ViT blocks. 
The decoder processes a combination of the encoded and masked patches and outputs the original reconstructed signal. A shared decoder is used to sequentially reconstruct each patch. %

\begin{table*}[!ht]
\centering
\small
\begin{tabularx}{\linewidth}{lXccccccc}
\toprule
Method & PT  & MT & \multicolumn{2}{c}{en-de $\downarrow$} & \multicolumn{2}{c}{en-it $\downarrow$} & \multicolumn{2}{c}{en-nl $\downarrow$} \\
 &  &  & Test-clean & Test-other & Test-clean & Test-other & Test-clean & Test-other \\
 \midrule
No Pre-training & None & - & 6.84  $\pm$ 0.22  & 12.91  $\pm$ 0.47 & 6.84  & 12.91  & 6.84  & 12.91  \\
\midrule
MAE & K600 & - & 7.54 & 13.88 & 7.54 & 13.88 & 7.54 & 13.88 \\
& K600 & \checkmark & \underline{5.69} & \underline{11.34} & \underline{5.95} & \underline{12.55} & \underline{5.73} & \underline{12.04} \\
& VC2 & - & 5.28 & 11.51 & 5.28 & 11.51 & 5.28 & 11.51 \\
& VC2 & \checkmark & \underline{5.11} & \underline{11.12} & \underline{5.56} & \underline{10.42} & 5.64 & 12.46 \\
& LRS3 & - & 4.73 & 10.27 & 4.73 & 10.27 & 4.73 & 10.27 \\
& LRS3 & \checkmark & 5.61 & 10.85 & \textbf{\underline{4.21}} & \textbf{\underline{9.53}} & 5.32 & 10.33 \\
\midrule
CLR & K600 & - & 6.85 & 12.92 & 6.85 & 12.92 & 6.85 & 12.92 \\
& K600 & \checkmark & \underline{5.02} & \underline{10.85} & \underline{4.72} & \underline{10.62} & \underline{4.65} & \underline{10.41} \\
& VC2 & - & 6.47 & 12.42 & 6.47 & 12.42 & 6.47 & 12.42 \\
& VC2 & \checkmark & \underline{6.43} & \underline{12.31} & \underline{5.1} & \underline{10.61} & \underline{4.62} & \underline{10.77} \\
& LRS3 & - & 6.35 & 12.12 & 6.35 & 12.12 & 6.35 & 12.12 \\
& LRS3 & \checkmark & 6.74 & \underline{10.59} & \underline{5.84} & \underline{11.33} & \underline{6.01} & \underline{10.13} \\
\midrule
MAE + CLR & K600 & - & 5.56 & 11.91 & 5.56 & 11.91 & 5.56 & 11.91 \\
& K600 & \checkmark & \underline{5.02} & \underline{11.68} & \underline{5.23} & \underline{11.37} & 6.39 & 12.03 \\
& VC2 & - & 6.75 & 12.11 & 6.75 & 12.11 & 6.75 & 12.11 \\
& VC2 & \checkmark & \underline{5.36} & \underline{11.22} & \underline{4.77} & \underline{10.84} & \underline{5.03} & \underline{10.73} \\
& LRS3 & - & 7.51 & 12.54 & 7.51 & 12.54 & 7.51 & 12.54 \\
& LRS3 & \checkmark & \underline{7.16} & \underline{12.29} & \underline{5.08} & \underline{11.13} & \underline{6.17} & \underline{12.32} \\
\bottomrule
\end{tabularx}
\caption{\small Performance (WER) on the Librispeech test-clean and test-other datasets with and without mid-training, and across Kinetics (K600), Voxceleb2 (VC2) and LRS-3 pre-training datasets. MT: With mid-training. MAE: Masked Autoencoding, CLR: Contrastive Learning. PT: Pre-Training. \underline{Underline} denote the consistent WER drop through mid-training alone across the 3 datasets and PT strategies. We observe that translation Mid-training task benefit the global representations of CLR more consistently compared to MAE. Overall, it improves the `only pre-trained' performance by aligning the learnt features towards the downstream task through auxiliary translation task. Further, italian language is the most effective as a mid-training task, suggesting that the languages that are complimentary to English may be more useful than others. }
\label{tab:main-table}
\end{table*}

\section{Results, Analysis \& Limitations}
\label{sec:results}

\begin{figure}[ht]
    \centering
    \includegraphics[width=0.87\linewidth]{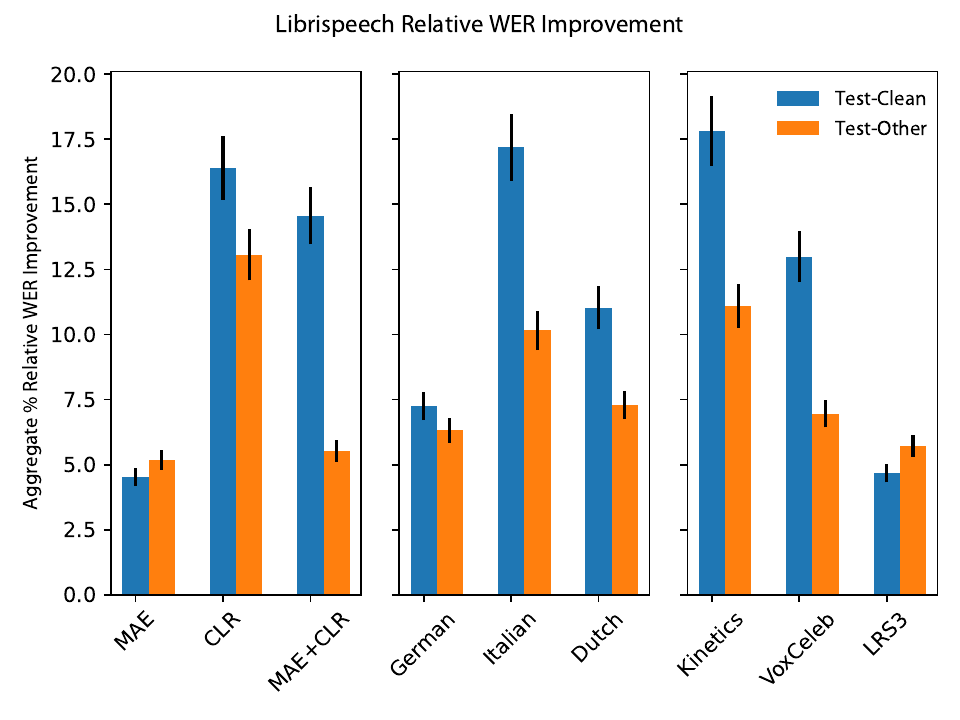}
    \caption{Average relative WER improvement on the Librispeech test-clean and test-other datasets with mid-training to show the effect of pre-training methods (left), mid-training translation pairs (center), and pre-training datasets (right). Translation mid-training improves upon CLR pre-training the most as it aligns its features for the local information required for ASR. Among the translation languages, Italian provides the best improvement, suggesting a complimentary language to English gains the most compared to languages that shares its roots with English. Models pre-trained on non-speech dataset Kinetics benefit the most from translation mid-training followed by noisy speech dataset Voxceleb2 and then clean speech dataset LRS3.  }
    \label{fig:lib}
\end{figure}

Our main results on the Librispeech dataset are shown in \autoref{tab:main-table} and \autoref{fig:lib}, and demonstrate several interesting learnings: \\

\noindent\textbf{Audio-visual Pre-training is Effective}: \autoref{tab:main-table} shows that on average in all cases, audio-visual pre-training is effective. Averaging the performance across all methods results in 6.34 $\pm$ 0.94 WER for test-clean, and  12.18 $\pm$ 0.98 for test-other. Under the null hypothesis that audio-visual pre-training is ineffective, we find significant improvements ($p = 0.035$) over the baseline. \\

\noindent\textbf{Mid-training with all translation pairs improve ASR performance}: \autoref{tab:main-table} shows that the mid-training approach leads to significant $(p < 0.01)$ improvements over pre-trained models alone, leading to relative WER improvements of 8.59\%/6.77\% (test-clean/test-other) with English-German pair,  18.55\%/10.28\% for English-Italian pair, and 13.11\%/7.71\% for English-Dutch pair. Surprisingly, Italian is the most effective, suggesting that choosing languages which are complementary to English may be more useful than languages which are closer to the target downstream language (English, Dutch and German all have Germanic roots, while Italian has Latin roots - see  \citet{tyshchenko2000metatheory} for a discussion on linguistic distance).

We leave it to future work to explore languages that retain very little shared information, such as Russian or Chinese.  The relative performance improvements with mid-training are shown in \autoref{fig:lib}. The figure shows several effects which we discuss in the following sections: the model pre-trained on Kinetics dataset is most improved with mid-training, English-Italian translation is the best mid-training  pair, and the model pre-trained with CLR benefits the most with mid-training. \\

\noindent\textbf{How do pre-training datasets impact performance (Is dataset size the only factor)?} Despite differences in pre-training dataset sizes, it is interesting to understand how the input mix of data impacts the overall performance of the model. Without mid-training, models pre-trained on LRS-3, the smalleset dataset, outperform all other models (6.19\%/11.64\% WER) on the test-other dataset. LRS-3 is a small fraction of the size of the VoxCeleb2 dataset, suggesting that the distributional makeup of the multi-modal dataset is key to pre-training performance, and \textit{dataset size is not all that matters}. VoxCeleb2 (6.16\%/12.01\% WER) outperforms LRS-3 slightly on the test-clean dataset. Kinetics trails both in aggregate (6.65\%/12.9\% WER), which could be due to both the size of the dataset (only half the size of VoxCeleb2), or the makeup of the dataset (no speech-specific data).

All three pre-training datasets outperform from scratch training for ASR (even Kinetics), indicating that \textit{pre-training on any amount or type of audio-visual data can be helpful}. We note that while Kinetics has the worst overall performance, it improves the most with mid-training (Rel. WER improvement of 14.03\%) vs VoxCeleb2 (9.45\%) and LRS-3 (6.17\%) (\autoref{fig:lib}). These results confirm that the model pre-trained on Kinetics has the most to gain from language-representation alignment (as it contains no speech data), and training on LRS-3, which consists of primarily clean data, has less to gain.

The best ASR results with MAE and CLR are obtained on the LRS-3 pre-training dataset. However the best MAE+CLR performance was in using the Kinetics dataset. While it can be difficult to disentangle the results from pre-training dataset size, this result may suggest that multi-task learning is more effective on out-of-domain data, where modalities contain non-redundant audio information, compared to VoxCeleb/LRS-3, where modalities consist of primarily redundant information. \\

\noindent\textbf{MAE outperforms CLR, MAE+CLR on ASR:} For ASR results averaged over all pre-training datasets, we find that MAE (5.63\%/11.53\% WER) alone outperforms both CLR (6.00\%/11.67\%) and MAE+CLR (6.09\%/11.85\%), suggesting that pre-training with masked auto-encoding objectives remains a promising approach for future exploration. Following intuition from \citet{chan2022avbert}, it is likely that CLR-augmented methods outperform on more global downstream tasks, whereas MAE encodes more local information which is useful for ASR, and MAE+CLR is a useful mix of both. This hypothesis is validated in our experiments on SUPERB \cite{yang2021superb}, where we found MAE+CLR most effective when aggregated across the mix of global (Intent Classification, Keyword Spotting), and local (Phoneme Recognition) tasks. \\

\begin{figure}[ht]
    \centering
    \includegraphics[width=\linewidth]{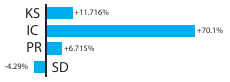}
    \vskip -0.1in
    \caption{\small Aggregate (dataset/language) relative performance improvement (higher is better) under mid-training for MAE + CLR on SUPERB. KS: Keyword spotting, IC: Intent Classification, PR: Phoneme Recognition, SD: Speaker Diarization. We observe consistent improvement in performance due to translation mid-training on tasks which require local feature information (KS, IC and PR) whereas global task SD observe a decrease in performance. It further shows that translation mid-training task enhances the pre-trained model's performance for local feature tasks while hurts the global feature task. }
    \label{fig:superb}
    \vspace{-1em}
\end{figure}

\noindent\textbf{Mid-Training is most effective with multi-task pre-training:} We explore the performance of our methods on four tasks from the SUPERB \cite{yang2021superb} benchmark in \autoref{fig:superb}. For SUPERB, mid-training improves performance for MAE+CLR models across most tasks. The notable exception is speaker diarization (SD), where there is minimal task overlap between SD and the mid-training target. Intent Classification (IC) is most improved (results not show in the tables), primarily due to a improvements in models pre-trained on the Kinetics (+80.17\%) and LRS-3 (+102.30\%) datasets, which benefit from the additional textual alignment. Keyword spotting (KS) improvements can also be largely attributed improvements on models pre-trained on Kinetics (+27.52\%), for similar reasons. Models pre-trained on VoxCeleb2 improve less with mid-training compared to models pre-trained with both Kinetics and LRS-3 for all tasks. We posit that since VoxCeleb2 dataset is already multi-lingual, and benefits less from further multi-lingual training. \\

\noindent\textbf{Note on baseline conformer performance:} In this work, we note that our baseline conformer models do not match the performance of \citet{gulati2020conformer}. Note that our primary goals was not to attain state of the art models, but study the impact of pre-training methods and datasets on ASR performance. The higher WER can be attributed to lower batch size used in out experiments, which was done to account for the large number of ablation studies done for this paper. While the overall baseline performance may be worse, the insights learned from the relative performance comparisons across the large-scale ablation are transferable to larger, more expensive models.\\

In summary, our results indicate the following:
\begin{itemize}
    \item Audio-visual pre-training is effective in almost all scenarios.
    \item Mid-training is useful and including data which is complementary is more effective than including data similar to pre-training data.
    \item Clean speech audio-visual dataset LRS-3 is an effective pre-training dataset given its size, compared to Kinetics and Voxceleb2.
    \item MAE pre-training is more effective than contrastive learning in ASR, while augmenting pre-training with CLR can help with downstream tasks that use global information.
\end{itemize}

\section{Discussion}

Recently, the size of pre-trained models and the datasets have increased to such an extent that it is cost-prohibitive to pre-train these models on datasets aligned with the downstream tasks of interest. Hence, a light-weight mid-training strategy can tune the pre-trained features strengthening the downstream performance.

 An alternative to the mid-training strategy is to include task during pre-training itself. This alternative strategy has two drawbacks; first, the amount of labeled data available for the mid-training task is typically not large enough to have significant impact when jointly learned in the pre-training stage. Secondly, the mid-training approach is more practical as it can be applied to already available pre-trained models instead of training the models from scratch which requires large amounts of time and compute.
\section{Conclusion \& Future Directions}

This work presents a multi-lingual mid-training objective and a large-scale analysis of multiple audio-visual pre-training methods and datasets, which confirms observations from \cite{hsu2021hubert} and \cite{chan2022avbert} --- we show how large scale audio-visual pre-training significantly improves downstream ASR performance, and that a well-chosen mid-training task can help the final downstream task.

While this paper presents initial insights into how mid-training tasks impact models multi-modal pre-trained models, we believe that significant additional future work remains to fully understand how sequences of training tasks can align large pre-trained models with downstream tasks. 

One interesting direction for future work is an exploration of additional mid-training tasks. In this work, we show that translation has the power to bridge gaps between multi-modal pre-trained models and language-based ASR tasks. Paired data for translation data can often be scarce, and may not be the optimal choice for future mid-training tasks. Instead, it may be insightful to explore mid-training tasks which are centered around synthetic data (such as TTS data from text datasets, or text generated by large language models) or self-supervised approaches to mid-training. 

Another closely related direction of future work explores how pre-training tasks impact the performance of downstream and mid-trained models. Here, we focus on multi-modal pre-training, as it is a key emergent direction of ASR research. However mid-training can easily be applied to uni-modal pre-training, or even zero-shot transfer from foundational models. 

In conclusion, this study sheds light on the impact of mid-training tasks in the context of multi-modal pre-training and demonstrates the significant improvement in downstream automatic speech recognition performance achieved through large-scale audio-visual pre-training. By continuing to delve into these areas, we can advance our understanding of how to effectively align pre-trained models with diverse downstream tasks and unlock new possibilities for multi-modal ASR research.

\nocite{*}
\section{Bibliographical References}\label{sec:reference}

\bibliographystyle{lrec-coling2024-natbib}
\bibliography{main}

\end{document}